\documentclass[letterpaper,10pt,conference]{IEEEtran}

\IEEEoverridecommandlockouts                              %

\usepackage{amsmath,amsfonts}
\usepackage{algorithmic}
\usepackage{algorithm}
\usepackage{array}
\usepackage[caption=false,font=normalsize,labelfont=sf,textfont=sf]{subfig}
\usepackage{textcomp}
\usepackage{stfloats}
\usepackage{url}
\usepackage{verbatim}
\usepackage{graphicx}
\usepackage{cite}
\usepackage[table]{xcolor}
\usepackage{booktabs}
\usepackage{hyperref}
\usepackage{multirow}
\usepackage{bm}
\title{\LARGE \bf
Learning Robot Policies for Whole-Body Mobility Assistance \\via Dynamic Human Motion Generation
}

\newcommand{\cmuaffiliation}{%
  \textit{The Robotics Institute,}\\
  \textit{School of Computer Science,}\\
  \textit{Carnegie Mellon University}\\
  Pittsburgh, USA%
}

\newcommand{\iaaffiliation}{%
  \textit{Intuitive Autonomy}\\
  San Jose, USA%
}

\makeatletter
\newcommand{\linebreakand}{%
  \end{@IEEEauthorhalign}
  \hfill\mbox{}\par
  \vspace{0.8em}
  \mbox{}\hfill
  \begin{@IEEEauthorhalign}
}
\makeatother

\author{%
  \IEEEauthorblockN{Shikun Ban\textsuperscript{*}}
  \IEEEauthorblockA{
    \cmuaffiliation\\
    sban@andrew.cmu.edu
  }
  \and
  \IEEEauthorblockN{Shohei Nagai\textsuperscript{*}}
  \IEEEauthorblockA{
    \cmuaffiliation\\
    shoheina@andrew.cmu.edu
  }
  \and
  \IEEEauthorblockN{Qingyang Song}
  \IEEEauthorblockA{
    \iaaffiliation\\
    oliver.song@\\
    intuitiveautonomy.com
  }

  \linebreakand
  
  \IEEEauthorblockN{Hanxi Wan}
  \IEEEauthorblockA{
    \iaaffiliation\\
    hanxi.wan@\\
    intuitiveautonomy.com
  }
  \and
  \IEEEauthorblockN{Eric Zhao}
  \IEEEauthorblockA{
    \iaaffiliation\\
    eric.zhao@\\
    intuitiveautonomy.com
  }
  \and
  \IEEEauthorblockN{Thomas Wang}
  \IEEEauthorblockA{
    \iaaffiliation\\
    tom.wang@\\
    intuitiveautonomy.com
  }
  \and
  \IEEEauthorblockN{Zackory Erickson}
  \IEEEauthorblockA{
    \cmuaffiliation\\
    zackory@cmu.edu
  }
  \thanks{*These authors contributed equally and are
  listed in alphabetical order.}%
}

\begin{document}

\maketitle
\thispagestyle{empty}
\pagestyle{empty}

\begin{abstract}
Robotic assistance for human mobility could benefit care recipients and reduce caregiver burden during physically demanding tasks such as Sit-to-Stand (STS) and Lift-and-Transfer (L\&T).
Learning robot policies for these whole-body assistive tasks is challenging: real-world data collection for human-robot interaction is costly and safety-critical, making simulation an attractive alternative. Prior work, however, typically treats the human as static or minimally responsive and thus fails to capture the cooperative nature of mobility assistance. We present a simulation-based framework for learning whole-body assistive robot policies. At its core is a human policy trained with an adversarial motion prior that produces realistic care-recipient motion in response to the robot's actions. A human-adaptive scripted controller interacts with this policy to collect diverse robot demonstrations at scale. From these demonstrations, we train 3D diffusion policies that map segmented human point clouds and robot joint positions to whole-body actions for a mobile manipulator, without relying on privileged human state at deployment. The resulting robot policies successfully perform both STS and L\&T assistance in simulation. Preliminary experiments further demonstrate sim-to-real transfer for STS after fine-tuning on a small amount of real-world data. Videos are available on our project page: \url{https://rchi-lab.github.io/human-gen-mobility-assistance/}
\end{abstract}

\begin{figure}[t]
    \centering
    \includegraphics[width=\columnwidth]{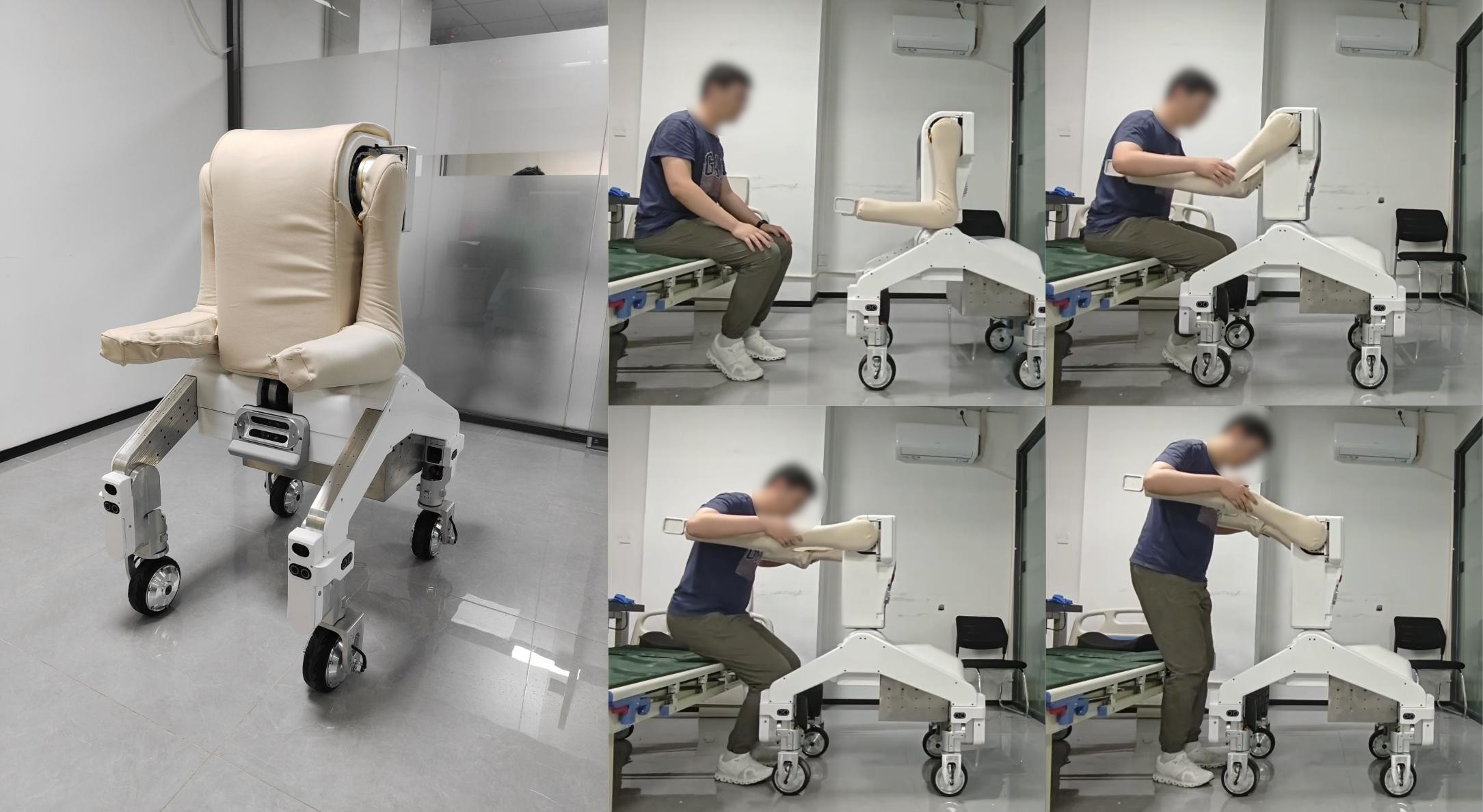}
    \caption{(Left) The bimanual mobile assistive robot platform used in this study. (Right) Sequence of STS assistance performed with the robot.}
    \label{fig:robot}
\end{figure}

\section{Introduction} 
Robotic assistance for human mobility can help individuals with severe motor impairments move through their environments and perform fundamental activities of daily living (ADLs).
Representative %
tasks include Sit-to-Stand (STS), where a caregiver helps the care recipient stand up from a seated pose, and Lift-and-Transfer (L\&T), where a caregiver supports, lifts, and transfers the care recipient from a seated pose to another location. 
Such assistance is frequent: a multicenter study found that $82.2\%$ of healthcare workers participated in patient-transfer tasks every working day, with $25.9\%$ performing such tasks at least six times per day~\cite{wahlin2025patient}.
However, these tasks impose substantial physical demands on caregivers.
They often require caregivers to support a large portion of the care recipient's body weight, resulting in repetitive joint loading that contributes to musculoskeletal injury and burnout~\cite{vinstrup2024ergonomic}.
These challenges motivate the development of physically assistive robots that can reduce caregiver burden while providing effective assistance (Fig.~\ref{fig:robot}).

A key challenge for such robots is deciding how to assist from observations of the human's motion and posture.
Human behavior during mobility assistance can vary substantially with physical condition, capability, and body morphology, creating a wide range of interaction patterns for the robot to handle.
This variability makes such tasks well suited to learning-based visuomotor policies that can adapt assistance behavior from onboard observations, following recent progress in training robot policies from large-scale imitation learning~\cite{chi2023diffusion_policy, kim2024openvla}.
However, learning robot policies for STS and L\&T assistance is challenging because these tasks involve tightly coupled physical interaction with a moving human, fundamentally different from many standard manipulation tasks involving passive objects.
Learning robust policies therefore requires interaction data that captures coupled dynamics between the robot and the human.
Collecting such data in the real world is prohibitively costly and poses safety concerns due to the close physical contact involved.
Although simulation offers a scalable alternative, many existing approaches to assistive robot policy learning rely on quasi-static or minimally active human models~\cite{Wang2023One, erickson2020assistive}.
These simplified models fail to capture how human motion evolves in response to robot assistance, limiting the diversity and physical fidelity of the resulting interactions.

In this work, we address this challenge by introducing a simulation-driven learning framework based on physics-based dynamic human motion generation.
Our key insight is that, because mobility assistance involves tightly coupled interaction with an actively cooperating human, the human must be modeled as an active agent whose motion reflects the variability of care recipients.
To this end, we train a physics-based human policy via reinforcement learning that produces biomechanically plausible human motions while interacting with the robot.
The human policy is trained with an Adversarial Motion Prior (AMP)~\cite{2021-TOG-AMP} using motion capture data, which encourages natural whole-body movement without requiring real-world human-robot interaction (HRI) demonstrations.
We further condition the human policy on a latent variable and optimize it with a diversity objective~\cite{eysenbach2018diversity} to capture different plausible ways of performing the same transfer task, such as variations in timing, posture, and interaction style.
As a result, the human policy produces diverse, physically realistic motions for mobility assistance.

We then leverage the active human model to collect robot demonstrations in simulation at scale.
Instead of relying on teleoperation, we generate demonstrations for the robot using a human-adaptive scripted controller with access to privileged human state.
Although the controller adapts to the current human state, the active human model exposes it to dynamically evolving human motions and physically coupled interaction patterns that are absent from quasi-static or replayed human models.
We train a 3D diffusion-policy for the robot that predicts delta joint commands from segmented human point clouds and robot joint positions, without privileged human state information.
Experiments show that our data-driven framework successfully learns whole-body assistive policies in simulation and supports preliminary real-world deployment.

Our main contributions are threefold:
\begin{enumerate}
    \item 
    We introduce the first policy-learning framework for STS and L\&T, two complex whole-body assistance tasks that support human mobility.

    \item We develop a dynamic physics-based human policy that models the care recipient as an actively cooperating agent during mobility assistance.

    \item We learn robot policies for both STS and L\&T using human-adaptive robot demonstrations collected in simulation, and demonstrate that the learned robot policies can provide successful mobility assistance and can be preliminarily deployed on a real-world assistive robotic system.
\end{enumerate}

\begin{figure*}[t]
    \centering
    \includegraphics[width=0.95\textwidth]{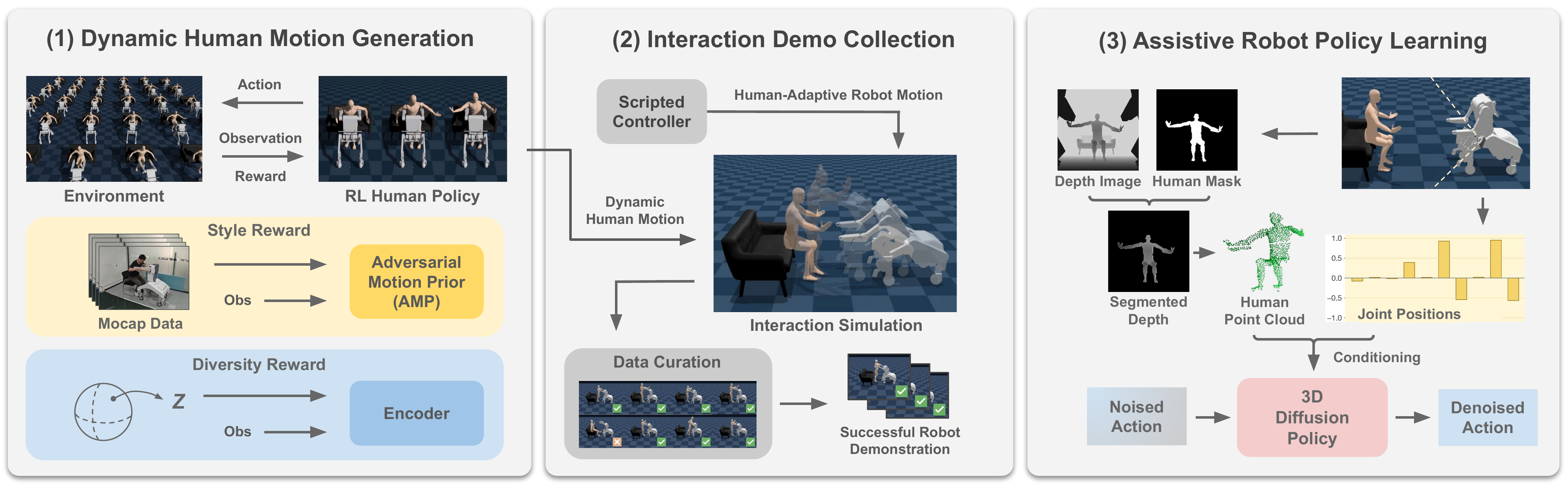}
    \caption{Our training pipeline consists of three stages. 
    (Left) We train a physics-based human policy via reinforcement learning, combining an Adversarial Motion Prior (AMP) for motion realism with latent conditioning to encourage diverse human behaviors. 
    (Middle) Using the trained human policy, we simulate robot assistance with a human-adaptive scripted controller. Successful interactions are filtered and stored as robot demonstrations. 
    (Right) We learn an assistive robot policy with a 3D diffusion-policy architecture using the collected demonstrations. At deployment, the policy predicts actions from human point clouds and robot joint positions without privileged human state information.}
    \label{fig:pipeline}
\end{figure*}

\section{Related Work}

\subsection{Assistive Robots for Dynamic Human Interaction}
Assistive robotics has been widely studied for activities of daily living, including feeding, dressing, and bathing~\cite{erickson2020assistive, 9982244, gordon2024adaptable, Wang2023One, madan2024rabbit, 11050927}.
These tasks typically involve static or minimally active care recipients, allowing policies to be trained with simplified human models.
In contrast, the STS and L\&T tasks studied in this work require dynamic, whole-body interaction in which the care recipient moves cooperatively with the robot.
A passive or oversimplified human model omits the dynamic human responses during deployment, creating a mismatch between simulated interactions and real-world human-robot interaction.
We close this gap by modeling the care recipient as an active agent, enabling robot policies to be trained under realistic, physically coupled HRI.

\subsection{Human Motion Generation}
Human motion generation has progressed from kinematic animation to physics-based methods that produce dynamically consistent and physically plausible behaviors~\cite{deepmimic2018, 2021-TOG-AMP, 2022-TOG-ASE, pan2024synthesizing}.
Prior work has also explored latent-conditioned policies to increase behavioral diversity across motion styles and skills~\cite{eysenbach2018diversity, 2022-TOG-ASE, luo2024universal}.
However, existing methods primarily target human-scene or human-object interaction, leaving physically coupled human-robot interaction largely unexplored.
Generating realistic motion in this setting requires a human model that is both biomechanically plausible and expressive enough to capture various care-recipient behaviors.
We address this gap by training a robot-conditioned human policy with AMP for biomechanical realism and latent conditioning to capture diverse, plausible behaviors, enabling realistic simulation of human motion during physical interaction with the robot.

\subsection{Vision-Based Robot Behavioral Cloning}

Behavioral cloning learns robot policies from expert observation--action pairs, with vision providing a scalable alternative when full-state information is unavailable.
Prior methods use RGB~\cite{chi2023diffusion_policy, mandlekar2021robomimic, shafiullah2022bet}, depth~\cite{patel2026defm, yuan2025depthvla}, or 3D representations~\cite{Ze2024DP3, gervet2023act3d, shridhar2023peract, 3d_diffuser_actor, goyal2023rvt, Wang2025articubot} to capture spatial structure for robot control.
Diffusion-based policies model multimodal action distributions from visual or 3D observations through conditional denoising~\cite{chi2023diffusion_policy, Ze2024DP3, ze2024humanoid_manipulation}.
For STS and L\&T, we use a human-centric 3D diffusion policy to capture spatial information during close interaction, mapping segmented human point clouds and robot joint positions to delta joint actions without privileged human state.

\section{Method}

\begin{figure}[t]
    \centering
    \includegraphics[width=1.0\linewidth]{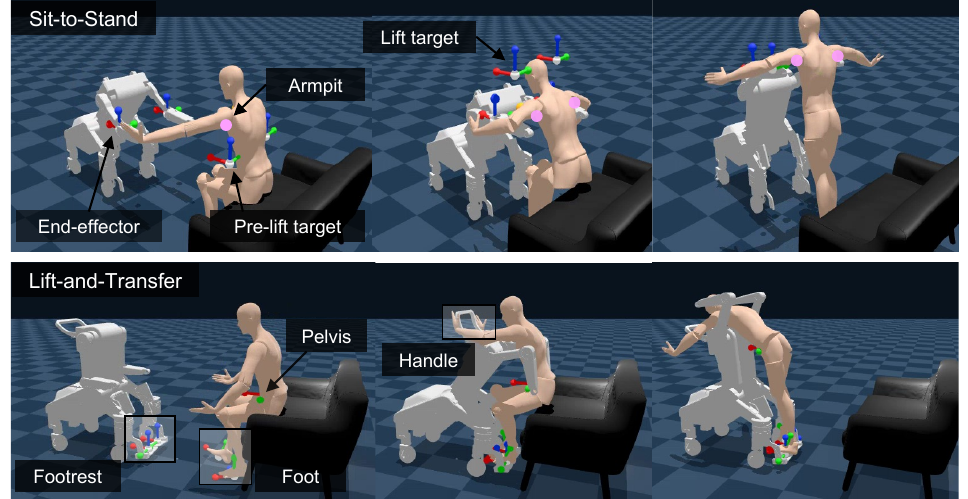}
    \caption{Task setups and key interaction points for scripted robot demonstration collection in STS (upper row) and L\&T (lower row).
    }
    \label{fig:interaction_link_kp}
\end{figure}

As shown in Fig.~\ref{fig:pipeline}, our proposed framework consists of three stages.
We first train an active physics-based human policy $\pi_{\mathrm{H}}$ to model dynamic care-recipient motion during mobility assistance.
We then use $\pi_{\mathrm{H}}$ with a human-adaptive scripted controller to collect successful robot demonstrations $\mathcal{D}$ in simulation.
Finally, we train a deployable assistive robot policy $\pi_{\mathrm{R},\theta}$ from $\mathcal{D}$, using only onboard observations.
The following sections describe each stage in detail.

\subsection{Task Setup}
\label{subsec:task_setup}
In this study, we instantiate our framework on two representative assistive tasks, performed with a bimanual mobile assistive robot platform (Fig.~\ref{fig:robot}) designed for close-contact mobility assistance.
The robot has a rear handle for the human to hold, footrests to stand on, and knee pads, enabling it to support and move the human after lifting.
Both tasks begin with the care recipient seated, and differ in how the robot supports and moves them (Fig.~\ref{fig:interaction_link_kp}):

\textbf{Sit-to-Stand (STS).} The care recipient rises from a chair to a standing position, while the robot positions its arms under the human's armpits and provides upward support.

\textbf{Lift-and-Transfer (L\&T).} The care recipient places their feet on the robot's footrests and grips a rear handle, while the robot positions its arms around the human's torso, lifts them up, and transfers them over a distance.

\subsection{Dynamic Human Motion Generation}
\label{subsec:human_motion_gen}

We train a physics-based human policy that generates dynamic human motion to simulate realistic care-recipient interaction during assistive tasks.
Realistic interaction requires motions that are physically plausible and span the wide distribution of real care-recipient behaviors, rather than following a single canonical trajectory.
To achieve this, we combine AMP~\cite{2021-TOG-AMP} for biomechanical realism with latent conditioning for diverse motion styles under the same task.

\subsubsection{State Representation}
The human policy observes an interaction state
\begin{equation}
s_t = (s_t^{\mathrm{H}}, s_t^{\mathrm{R}}, s_t^{\mathrm{E}}),
\end{equation}
where $s_t^{\mathrm{H}}$ denotes the human state, $s_t^{\mathrm{R}}$ denotes task-relevant robot states, and $s_t^{\mathrm{E}}$ denotes environment states.
Following prior work~\cite{hassan2023synthesizing}, $s_t^{\mathrm{H}}$ includes root height, root linear and angular velocities, local joint rotations, local joint velocities, and key body link positions: hands, feet, and head.
The robot state $s_t^{\mathrm{R}}$ contains task-specific robot link positions used to model physical interaction, while $s_t^{\mathrm{E}}$ contains the chair position.

\subsubsection{Realism via Adversarial Motion Prior}
To encourage realistic motion, we use an AMP~\cite{2021-TOG-AMP} where the discriminator $D(s_t^{\mathrm{H}}, s_{t+1}^{\mathrm{H}})$ is trained to distinguish simulated human state transitions from motion capture transitions, and the human policy receives the style reward:
\begin{equation}
r_t^{\mathrm{S}}
=
-\log \left(1 - D(s_t^{\mathrm{H}}, s_{t+1}^{\mathrm{H}})\right).
\end{equation}
The discriminator is defined only over human states, encouraging natural whole-body movement without requiring motion capture of robot or chair poses, which are difficult to obtain reliably during close physical interaction.

\subsubsection{Diversity via Latent Conditioning}
Care-recipient motion can vary across individuals and trials in motion timing, posture, and interaction style during the same task.
To capture this variability, we condition the human policy on an episode-level latent code $z$ sampled once per episode.
Following~\cite{2022-TOG-ASE}, we draw $\tilde{z} \sim \mathcal{N}(0, I)$ and project it onto the unit hypersphere, $z = \tilde{z} / \|\tilde{z}\|_2$:
\begin{equation}
a_t^{\mathrm{H}} \sim \pi_{\mathrm{H}}(\cdot \mid s_t, z).
\end{equation}
Following latent skill discovery methods~\cite{eysenbach2018diversity, 2022-TOG-ASE}, we maximize the mutual information between $z$ and the resulting interaction transition using a variational encoder $q_\phi(z \mid s_t, s_{t+1})$.
This yields the diversity reward
\begin{equation}
r_t^{\mathrm{D}}
=
\log q_\phi(z \mid s_t, s_{t+1}).
\end{equation}
Because the encoder observes the full interaction transition, including human, robot, and environment states, the latent variable encourages diverse interaction behaviors rather than isolated kinematic variations in human motion.
In contrast to prior work where latents often represent distinct skills, we use $z$ to represent different plausible human motions of performing the same assistive task.

\subsubsection{Human Policy Objective}
The human policy is trained with the total reward
\begin{equation}
r_t
=
r_t^{\mathrm{S}}
+
w^{\mathrm{D}} r_t^{\mathrm{D}}
+
w^{\mathrm{G}} r_t^{\mathrm{G}},
\end{equation}
where $r_t^{\mathrm{S}}$ encourages motion realism, $r_t^{\mathrm{D}}$ encourages latent-conditioned behavioral diversity, and $r_t^{\mathrm{G}}$ is a task-specific goal reward that guides the human toward successful task participation.
For STS, $r_t^{\mathrm{G}}$ includes terms for head height, foot--ground contact, arm--robot contact, and armpit proximity to the robot arms.
For L\&T, $r_t^{\mathrm{G}}$ includes terms for handle proximity, chest contact, footrest proximity and contact, and knee-pad proximity.
Both tasks also apply a small penalty on squared joint accelerations to encourage smooth motion.

\subsubsection{Motion Capture Dataset}
We collect task-specific reference motions for the AMP discriminator using an OptiTrack motion-capture system.
The dataset includes twelve demonstrations per task from six participants, recorded at 120~Hz and lasting approximately 10--20~s each.
For STS, participants interact with a second person emulating the robot arms with two sticks; for L\&T, participants interact with the real robot executing a scripted motion.
The captured trajectories are then manually retargeted onto the 31-DoF skeleton~\cite{alhafez2023b}.

\subsubsection{Training}
We train the human policy with PPO~\cite{schulman2017proximal} in the Genesis simulator~\cite{Genesis} using a 31-DoF articulated human skeleton~\cite{alhafez2023b} with a human skin mesh~\cite{briceno2018makehuman} for depth rendering.
During human policy training, the robot follows scripted task motions, which exposes the human policy to task-relevant contact and support dynamics.
To prevent the human policy from overfitting to a single robot behavior or interaction pattern, at each episode reset we uniformly perturb the initial human and robot positions ($\pm0.10$~m), yaw ($\pm\pi/18$~rad), and human joint angles ($\pm\pi/18$~rad) about their defaults, and sample the chair height from $0.40$~m to $0.51$~m.

After training, the human policy is frozen and used as an active simulated care recipient for robot demonstration collection.
The frozen human policy is intended to model task-relevant responses to assistive robot motions, rather than recovery from arbitrary out-of-distribution robot behaviors; such interactions are filtered during demonstration collection and treated as task failures during evaluation.

\subsection{Robot Interaction Demonstration Collection}
\label{subsec:demo_collection}

To collect robot demonstrations, we deploy a human-adaptive scripted controller in simulation together with the frozen human policy 
$\pi_{\mathrm{H}}$.
Fig.~\ref{fig:interaction_link_kp} illustrates the task setups and key interaction points used by the scripted controller.

For STS, the controller tracks three sequential Cartesian end-effector targets defined relative to the current human armpit positions and root yaw: reaching below the armpits, lifting to a body-scale-dependent height, and slowly retreating from the human.
At each time step, the next end-effector target is converted to desired robot joint positions using inverse kinematics.
For L\&T, the controller follows a scripted whole-body joint-position trajectory conditioned on the human state: the robot base is shifted toward the sampled human position, and the support/lifting motion is scaled with body size to avoid over- or under-extension.
In both tasks, desired joint positions are updated incrementally and tracked at $100$ Hz using joint-space PD control subject to actuator effort limits, producing smooth robot motion during contact.

During demonstration collection, we randomize human physical capability, body scale, motion latent code, initial body joints, relative human-robot placement, and furniture height.
Human capability is modeled by scaling the baseline joint torque limits of the simulated human model.
Human body scale is sampled from $\{0.88, 1.0\}$, and the 4-dimensional motion latent code is sampled once per episode.
We run the scripted controller over multiple episodes and retain successful trajectories 
$\tau_i = \{(o_t^{\mathrm{R}}, a_t^{\mathrm{R}})\}_{t=1}^{T_i}$,
where $o_t^{\mathrm{R}}$ and $a_t^{\mathrm{R}}$ denote the robot observation and action at time step $t$.
After filtering, we obtain a robot demonstration dataset 
$\mathcal{D} = \{\tau_i\}_{i=1}^N$ 
covering robot actions that adapt to diverse human motions and interaction configurations.

\subsection{Assistive Robot Policy Learning}

Given the curated demonstration dataset $\mathcal{D}$, we learn an assistive robot policy 
$\pi_{\mathrm{R},\theta}: \mathcal{O}^{\mathrm{R}} \mapsto \mathcal{A}^{\mathrm{R}}$ 
that maps onboard observations to robot actions 
(Fig.~\ref{fig:pipeline} (Right)). 

\subsubsection{Observation and Action Space}
Each observation 
$o_t^{\mathrm{R}}$
consists of a human point cloud 
$p_t^{\mathrm{H}}$, obtained by applying a simulator-provided human segmentation mask to the depth observation and back-projecting the remaining depth pixels, together with robot joint positions $q_t$.
Each action 
$a_t^{\mathrm{R}}$
is represented as a sequence of robot delta joint commands.

\subsubsection{3D Diffusion Policy Formulation}
We train the policy with behavioral cloning using a conditional 3D diffusion policy~\cite{Ze2024DP3} that models $p(a_t^{\mathrm{R}} \mid o_t^{\mathrm{R}})$.
The human point cloud is encoded using a PointNet-based encoder~\cite{qi2017pointnetplusplus}. 
The resulting point-cloud feature is concatenated with the robot joint positions $q_t$ and used as the conditioning input to the diffusion model.
Starting from Gaussian noise $a_t^{\mathrm{R},K}$, the denoising network $\epsilon_\theta$ iteratively refines the action sequence:
\begin{equation}
a_t^{\mathrm{R},k-1} = \alpha_k \left(
a_t^{\mathrm{R},k} - \gamma_k \epsilon_\theta(a_t^{\mathrm{R},k}, k, o_t^{\mathrm{R}})
\right) + \sigma_k \mathcal{N}(0, I),
\end{equation}
where $\alpha_k, \gamma_k, \sigma_k$ are determined by the squared cosine noise schedule from iDDPM~\cite{nichol2021improved}.

\subsubsection{Training}
During training, a ground-truth delta joint action sequence $a_t^{\mathrm{R},0}$ is corrupted with Gaussian noise at diffusion time step $k$:
\begin{equation}
a_t^{\mathrm{R},k} = \sqrt{\bar{\alpha}_k}a_t^{\mathrm{R},0} + \sqrt{1-\bar{\alpha}_k}\epsilon^k,
\end{equation}
where $\epsilon^k \sim \mathcal{N}(0, I)$. 
The denoising network is trained to predict the injected noise:
\begin{equation}
\mathcal{L}
=
\mathrm{MSE}\big(
\epsilon^k,
\epsilon_\theta(a_t^{\mathrm{R},k}, k, o_t^{\mathrm{R}})
\big).
\end{equation}

\subsubsection{Inference}
At inference time, the denoised action sequence is executed using receding-horizon control, where only the first few predicted actions are applied before replanning from the latest observation. 
Each predicted action is a delta joint command $\Delta q_t$, and the target joint position is computed as 
$q_{t}^{\mathrm{target}} = q_{t} + \Delta q_t$.
Because the policy uses only segmented human point clouds and robot joint positions, it can be deployed with visual and proprioceptive observations alone, without requiring access to the privileged human state used during demonstration collection.

\section{Experiment}
\label{sec:experiment}
Our experiments address three research questions:
(1) Can the proposed human policy produce realistic and diverse motions during physical HRI?
(2) Can the robot policy learn robust assistive behavior from demonstrations collected with these human motions?
(3) Can the resulting robot policy transfer to a real-world assistive robot?
The study covers the two assistive tasks described in Sec.~\ref{subsec:task_setup}: \textbf{STS} and \textbf{L\&T}.
We use a bimanual mobile assistive robot with 18 controllable DoFs: 3 chassis, 8 lower-body including 4 wheel joints, 1 chest-bending, and 6 arm joints.
Two front-facing RGB-D cameras provide $1280 \times 720$ images with an approximately $60^\circ$ field of view each, together covering about $120^\circ$.
The simulated robot matches the controllable joint structure of the physical platform.

\subsection{Human Motion Generation Evaluation}
We first evaluate whether the proposed human policy generates motions that are both realistic and diverse under physical interaction with the robot.
A usable human model must balance these two properties: motions should remain realistic enough to reflect plausible human behavior, while also being diverse enough to provide varied interaction data for robust robot policy learning.
To isolate the contribution of each component, we conduct an ablation study with four variants:
\begin{enumerate}
\item \textit{AMP+Latent (Ours)}: Trained with both AMP and latent conditioning.
\item \textit{AMP-only}: Trained with AMP only.
\item \textit{Latent-only}: Trained with latent conditioning only.
\item \textit{Vanilla}: Trained without AMP or latent conditioning.
\end{enumerate}

For the latent-conditioned variants, a 4D latent code is sampled from a standard normal distribution at the beginning of each episode and projected onto the unit hypersphere.
Since non-AMP variants do not reliably complete L\&T, we conduct this evaluation only on STS.

\begin{table}[t]
\centering
\caption{Evaluation of generated human motions on STS.}
\label{tab:human_motion_eval}
\begin{tabular}{lcc}
\hline
Variant & Disc. Score ($\uparrow$) & Joint Angle Std. ($\uparrow$) [$\mathrm{rad}$] \\
\hline
AMP+Latent (Ours) & $\mathbf{0.384 \pm 0.052}$ & \underline{0.091} \\
AMP-only          & \underline{0.356 $\pm$ 0.053}       & 0.030 \\
Latent-only       & $0.002 \pm 0.010$                   & $\mathbf{0.141}$ \\
Vanilla           & $0.003 \pm 0.008$                   & 0.048 \\
\hline
\end{tabular}
\vspace{-0.4cm}
\end{table}

\subsubsection{Metrics}
We evaluate generated human motions along two criteria: realism and diversity.

\textbf{Realism.}
We measure realism with an AMP-style discriminator $D(s,s')$, trained independently per variant to distinguish generated state transitions from motion-capture transitions. We report the mean discriminator score over all generated transitions, where a higher score indicates motion that is more consistent with the motion-capture distribution.

\textbf{Diversity.}
We measure diversity using the joint angle standard deviation $\sigma_{\text{joint}}$, which quantifies how much generated human motions vary across episodes:
\begin{equation}
\sigma_{\text{joint}} =
\sqrt{
\frac{1}{TNJ}
\sum_{t=1}^{T}
\sum_{i=1}^{N}
\left\| q^{(i)}_t - \bar{q}_t \right\|_2^2
},
\end{equation}
where $T$ is the number of time steps, $N$ is the number of episodes, $J$ is the number of joints, $q^{(i)}_t$ is the joint angle vector at time $t$ in episode $i$, and $\bar{q}_t$ is the mean joint angle vector at time $t$ averaged across episodes.

\subsubsection{Quantitative Results}
Table~\ref{tab:human_motion_eval} reports the discriminator score and joint angle standard deviation for all variants.
The results show that AMP is essential for realistic human motion generation.
\textit{AMP+Latent (Ours)} achieves the highest discriminator score, while \textit{AMP-only} obtains the second-highest score.
In contrast, both non-AMP variants score near zero.

The diversity results show the complementary role of latent conditioning.
\textit{AMP-only} produces realistic motions, but has the lowest joint angle standard deviation, suggesting that it generates a relatively narrow set of motions under the same task condition.
\textit{Latent-only} achieves the highest diversity, but this diversity comes from unrealistic motions, as reflected by its near-zero discriminator score.
Our method achieves the second-highest diversity while maintaining the highest realism score, suggesting AMP and latent conditioning provide a practical balance between realism and behavioral diversity.

\subsubsection{Qualitative Results}
\begin{figure}[t]
    \centering
    \includegraphics[width=\columnwidth]{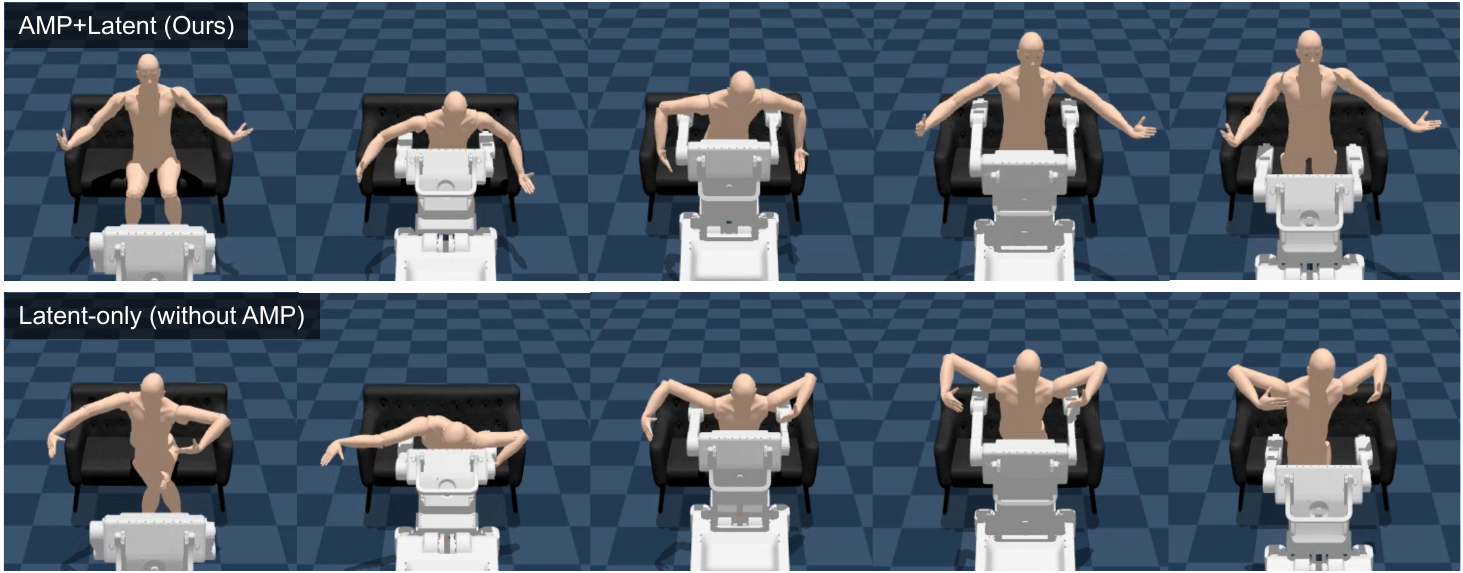}
    \caption{Qualitative comparison of generated STS motions. Top: AMP-based variants generate realistic interaction motions. Bottom: Non-AMP variants exhibit unnatural postures.}
    \label{fig:human_motion_qualitative}
\end{figure}

Fig.~\ref{fig:human_motion_qualitative} qualitatively contrasts our method with a non-AMP variant (\textit{Latent-only}).
The AMP-based variants produce interaction motions that visually resemble real STS, whereas the non-AMP variants exhibit reward-hacking artifacts and unnatural postures that break realistic human-robot coordination.
These qualitative results are consistent with the discriminator scores and further support the importance of AMP for motion realism.

Overall, these results show that the proposed method equips the human model with realistic yet diverse motions.
In the subsequent robot policy evaluation, we therefore use this variant as the human model.

\subsection{Robot Policy Evaluation in Simulation}

For STS and L\&T, we train separate robot policies using 3D Diffusion Policy~\cite{Ze2024DP3} with point-cloud observations encoding the human geometry and relevant spatial configuration. Other architectures could also be used in principle. We down-sample each point cloud to 1024 points via farthest point sampling, use delta joint-position actions, and set prediction and execution horizons to 16 and 8, respectively.
We construct the training datasets from successful scripted-controller demonstrations (Sec.~\ref{subsec:demo_collection}), adding data for underrepresented capability and body-scale settings to reduce bias. The resulting STS and L\&T datasets contain 957 and 589 demonstrations, totaling 539,664 and 1,706,497 frames, respectively, with the larger L\&T frame count due to longer episodes. On a single NVIDIA RTX 4090, we train the STS policy for approximately 1.3M gradient updates and the L\&T policy for 2.0M. Next, we present the detailed metrics and results of our robot policy evaluation.

\subsubsection{Metrics}
\label{sec:metric}

We evaluate the learned robot policy over five groups of 60 simulation cases for each task, and report the mean and standard deviation across groups. 
Evaluation cases are sampled separately from those used for demonstration collection.
We report two metrics: \textbf{Pre-lift SR} (success rate) and \textbf{Final SR}.   
Pre-lift success measures whether the robot reaches a valid assistance configuration prior to lifting, while final success measures whether the full assistive task is completed. 

For STS, pre-lift success requires each robot end-effector to be within $0.10$ m of the corresponding human upper-arm centroid along the robot arm direction.
The human elbows should lie laterally outside the corresponding robot arms, and the end-effectors should remain below the paired armpits.
Together, these criteria indicate that the robot arms are positioned beneath the human armpits and near the human upper arms, ready to provide lifting support.
Final success requires the head to exceed a scale-adjusted height threshold ($1.58\,\mathrm{m}\times$ human scale) and the hip flexion angle to satisfy an upright-posture threshold ($<0.65$ rad).
Together, these criteria indicate sufficient lift height and lower-body extension rather than a seated-like posture.

For L\&T, Pre-lift success requires sufficient footrest contact before lifting, defined by the maximum summed vertical contact force between both feet and footrests exceeding $400\,\mathrm{N}\times$ human scale, where human scale approximates body-mass variation.
Final success requires the shoulders to rise above the robot arms, the pelvis to pass a distance threshold, and the summed footrest contact force to exceed $300\,\mathrm{N}\times$ human scale.
Together, these criteria indicate successful lifting and transfer.

\subsubsection{Task Performance}
\label{sec:task_performance}

\begin{table}[t]
\centering
\caption{Robot policy performance in simulation.}
\label{tab:sim_results}
\resizebox{0.8\columnwidth}{!}{%
\begin{tabular}{lcc}
\hline
Task & Pre-lift SR (\%, $\uparrow$) & Final SR (\%, $\uparrow$) \\
\hline
STS & $98.3 \pm 2.4$ & $95.7 \pm 2.7$ \\
L\&T & $90.0 \pm 6.7$ & $74.0 \pm 9.0$ \\
\hline
\end{tabular}}
\end{table}

\begin{figure}[t]
    \centering
    \includegraphics[width=\columnwidth]{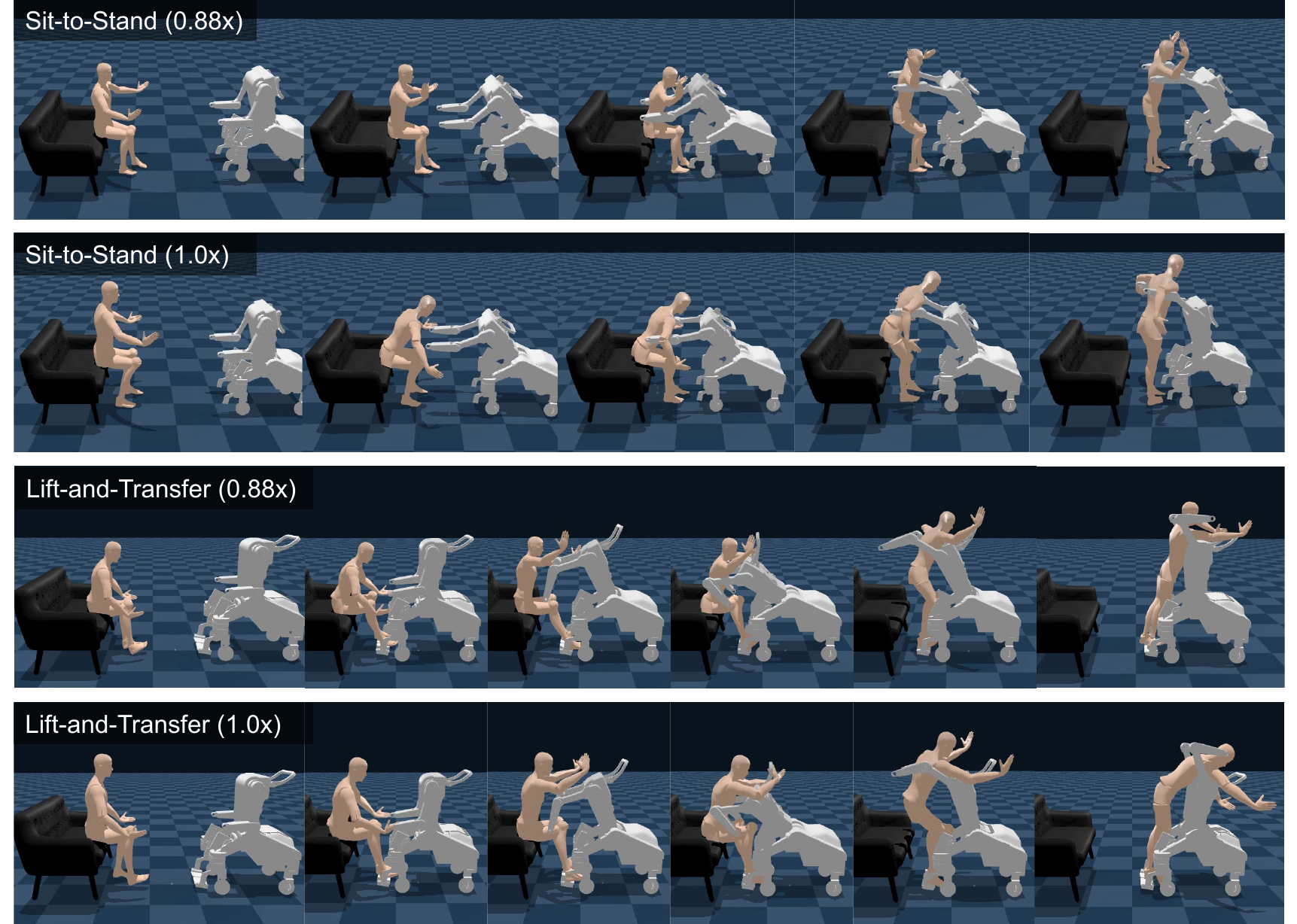}
    \caption{
Representative successful evaluation rollouts of our learned policies for STS and L\&T.
For each task, the upper and lower sequences correspond to human body scales of $0.88\times$ and $1.0\times$, respectively.
The robot reaches the pre-lift configuration and completes the assistance in all examples.
}
    \label{fig:rollout_vis}
\end{figure}

As shown in Table~\ref{tab:sim_results}, the policy achieves relatively high Pre-lift SRs of $98.3\%$ on STS and $90.0\%$ on L\&T.
It achieves a $95.7\%$ Final SR on STS and a $74.0\%$ Final SR on L\&T. Fig.~\ref{fig:rollout_vis} shows representative successful evaluation rollouts of the learned policy for both tasks under different human body scales. These examples illustrate that our policy can adapt robot arm positioning and lifting height to different human scales when completing the assistance sequence.

\begin{figure}[t]
    \centering
    \includegraphics[width=\columnwidth]{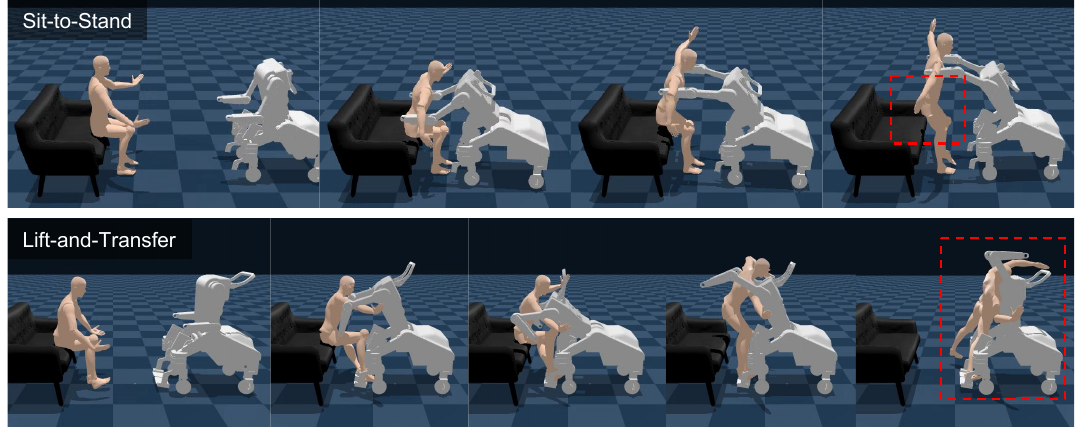}
    \caption{
    Representative failure rollouts of our learned policies.
Top: in STS, the robot does not lift the human sufficiently high, leaving the human in a non-upright posture.
Bottom: in L\&T, the robot places its arms incorrectly, preventing the human from reaching the handle and being properly lifted.
Red dashed boxes highlight the main failure regions.
    }
    \label{fig:failure_examples}
\end{figure}

Fig.~\ref{fig:failure_examples} visualizes representative failure modes.
In STS, the robot does not lift the human sufficiently high, leaving the human below the head-height threshold and in a non-upright hip posture.
Nevertheless, the robot still makes meaningful progress toward standing.
In L\&T, incorrect robot arm placement can disrupt coordination, preventing the human from reaching the handle and being properly lifted.
These failures may stem from partial human occlusion that degrades point-cloud observations, and limited recovery examples in the demonstrations for correcting early errors.

\subsubsection{Ablation on Human Capability in Robot Policy Training}
\label{sec:ablation_human_capability}

\begin{table}[t]
\centering
\caption{Ablation on human capability in STS robot policy training.}
\label{tab:ablation_human_capability}
\resizebox{1.0\columnwidth}{!}{%
\begin{tabular}{lcc}
\hline
Robot Policy Trained With & Pre-lift SR (\%, $\uparrow$) & Final SR (\%, $\uparrow$) \\
\hline
Mixed $1.0\times, 0.25\times$ capability (Ours) & $\mathbf{98.3 \pm 2.4}$ & $\mathbf{95.7 \pm 2.7}$ \\
Single $1.0\times$ capability & $\underline{96.7 \pm 2.7}$ & $90.3 \pm 7.1$ \\
Single $0.25\times$ capability & $\mathbf{98.3 \pm 2.4}$ & $\underline{94.0 \pm 4.4}$ \\[1.5pt]
\hline
\end{tabular}}
\end{table}

We conduct this ablation on STS to examine whether our mixed-capability training strategy enables more reliable assistance across human capability levels.
Human capability is modeled by scaling the articulated human's baseline joint torque limits.
We compare our method, training robot policy with demonstrations from both $1.0\times$ and $0.25\times$ capability humans, against two single-capability baselines: Single $1.0\times$ capability and Single $0.25\times$ capability.
For fair comparison, all policies are trained with the same number of demonstrations and gradient updates. They are evaluated on the same capability sweep of $1.0\times$, $0.75\times$, $0.5\times$, and $0.25\times$, including intermediate levels not used during training.

As shown in Table~\ref{tab:ablation_human_capability}, all policies achieve high pre-lift success rates.
Our mixed-capability-trained policy achieves the highest Final SR, with a clear improvement over the $1.0\times$-only baseline and a smaller gain over the $0.25\times$-only baseline.
These results suggest that training with varied human capabilities improves robustness across capability variation, a critical requirement for assistive robots providing support for users with different physical abilities.
The stronger performance of the $0.25\times$-only baseline over the $1.0\times$-only baseline further indicates that lower-capability humans may expose the policy to more demanding assistance dynamics and improve policy robustness.

\subsubsection{Verifying Robot Assistance Necessity}

\begin{figure}[t]
    \centering
    \includegraphics[width=\linewidth]{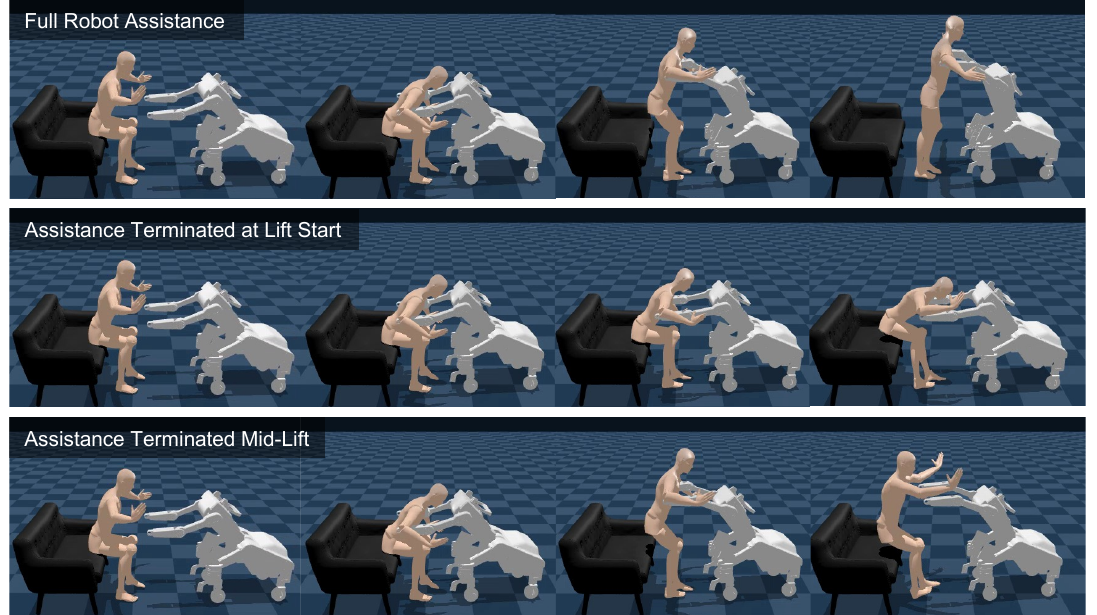}
\caption{
The learned human models are not able to stand up independently.
The top row shows a successful rollout with full robot assistance.
The middle and bottom rows show the same rollout with assistance terminated at lift start and mid-lift, respectively.
In both terminated conditions, the human fails to stand without sufficient robot assistance.
}
    \label{fig:sanity_check}
\end{figure}

We verify that the learned human policy cannot complete the task without sufficient robot assistance (i.e. not able to stand independently).
Quantitatively, we randomly select 60 successful STS evaluation cases, with 30 at each human body scale ($0.88\times$ and $1.0\times$), and rerun them with the human policy unchanged.
To isolate continued robot assistance, the robot follows the original lifting motion until its end-effectors reach a height of $1.00$ m, then stops lifting before completing the assistance.
By this point, the human has been lifted by an average of $0.15$ m from the seated position.
None of these early-terminated rollouts succeeds ($0/60$), showing that STS completion requires sufficient robot assistance.
Qualitatively, Fig.~\ref{fig:sanity_check} compares full robot assistance with assistance terminated at lift start and mid-lift.
Although the human is lifted off the seat, it fails to complete standing after robot assistance terminates.

\subsection{Real-World Robot Deployment}

\begin{figure*}[t]
    \centering
    \includegraphics[width=1.00\textwidth]{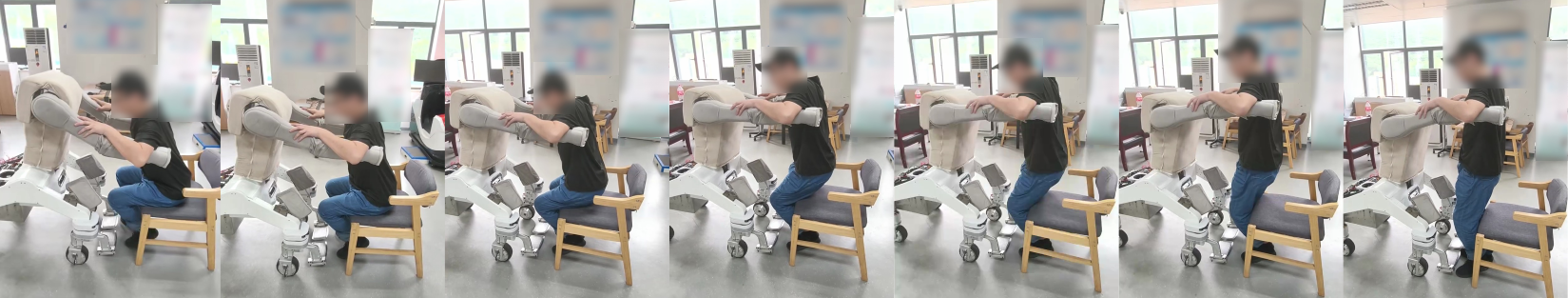}
    \caption{Real-world deployment of the learned assistive robot policy. The fine-tuned policy assists a human participant in a real-world STS interaction.
    }
    \label{fig:real_world_demo}
\end{figure*}

To evaluate sim-to-real transfer, we deploy the learned policy on a physical assistive robot, as shown in Fig.~\ref{fig:real_world_demo}. 
For safety in this initial study, the final deployment is conducted with one able-bodied human participant.
Because STS requires close under-arm contact, we first test the simulation-trained policy on a manikin rather than directly running trials with the participant.
This diagnostic test suggested that real-world policy fine-tuning would be necessary to achieve sim-to-real transfer of the robot policy as the robot arms moved upward toward the support region but did not reliably reach the target height needed for standing assistance.

Since real-human deployment further increases the gap from simulation in body shape, motion, and compliant contact, we perform lightweight fine-tuning from the simulation-trained policy using five real-world STS robot demonstrations, totaling $385.64$ s ($6$ min $26$ s) of data.
These demonstrations are collected with one assistance recipient and one robot operator using a state-machine STS controller.
The controller segments the human, detects armpit keypoints, and plans robot arm motions for STS assistance.
The deployment participant is not included in either the motion-capture dataset used to train the human motion prior or the real-world demonstrations used for fine-tuning.
The full policy is fine-tuned using the same diffusion objective for $60{,}000$ gradient updates (less than $5\%$ of the simulation-training budget), with observation augmentation to account for real-world sensing noise and body-shape variation.
This setup evaluates whether the simulation-trained policy provides a useful initialization for adaptation with limited real-world data.

During real-world deployment, YOLO-26 segments the human in onboard RGB-D images, and the segmented depth pixels are back-projected to form the human point-cloud input; as in simulation, the policy outputs delta joint-position commands. For safety, commands are clipped to position and velocity limits, the robot operates at reduced speed, and an operator supervises each rollout with access to an emergency stop.
As shown in Fig.~\ref{fig:real_world_demo}, the fine-tuned policy performs STS assistance on the physical robot, providing initial evidence that our simulation-trained policy can support real-world physical assistance after lightweight fine-tuning.

\section{Limitations and Future Work}

This work has several limitations that motivate future work.
First, real-world evaluation remains limited to STS with one able-bodied participant; future work should extend it to diverse participants, including care recipients with mobility impairments.
Second, the current system does not directly regulate assistive forces; incorporating force sensing and human feedback could help modulate support during lifting.
Third, simulated humans still differ from real users and target patient populations. Additional motion-capture data, human video priors, and musculoskeletal modeling could reduce this gap and improve the reliability of sim-to-real deployment.

\section{Conclusion}

This work introduces a scalable, simulation-driven framework for learning robust robot policies capable of executing complex, whole-body assistive tasks, including Sit-to-Stand (STS) and Lift-and-Transfer (L\&T). To overcome the data collection and safety challenges inherent in close-contact human-robot interaction, we develop a physics-based human policy driven by an adversarial motion prior and latent conditioning. This approach successfully models the care recipient as an actively cooperative agent, yielding realistic and diverse human interaction dynamics. By using this model to generate simulated demonstrations, we train point-cloud-conditioned 3D diffusion policies for STS and L\&T that achieve high success rates in simulation and exhibit robustness to varying human physical capabilities. Furthermore, we demonstrate the real-world viability of our method by successfully deploying the STS policy on a physical bimanual assistive robot following lightweight fine-tuning.

\section*{Acknowledgments}
This work was supported by Intuitive Autonomy. We thank the technical team and study participants at Intuitive Autonomy for their support with simulation development, motion-capture data collection, and real-world robot testing. We also thank Kavya Puthuveetil and Yucheng Li for reviewing the manuscript and providing valuable feedback.

\bibliographystyle{IEEEtran}
\bibliography{root}

\vfill

\end{document}